\documentclass[conference]{IEEEtran}

\usepackage{graphicx}
\usepackage{amsmath}
\usepackage{url}
\usepackage[bookmarks=false]{hyperref}
\usepackage{subcaption}
\usepackage{cuted}
\usepackage{capt-of}
\usepackage{flushend}

\graphicspath{{}}

\IEEEoverridecommandlockouts
\begin{document}

\title{Texel Splatting: Perspective-Stable 3D Pixel Art}

\author{
\IEEEauthorblockN{Dylan Ebert}
\thanks{Code and demo: \url{https://dylanebert.com/texel-splatting}}
}

\maketitle

\begin{strip}
\centering
\vspace{0.5em}
\includegraphics[width=\linewidth]{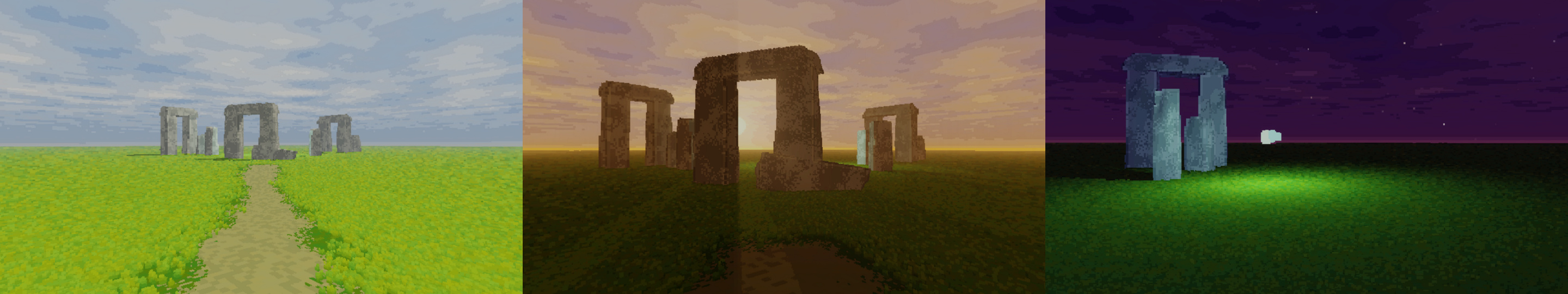}
\captionof{figure}{Texel splatting renders 3D scenes as perspective-stable pixel art. Each cubemap texel is splatted to screen as a world-space quad that remains fixed to geometry as the camera moves.}
\label{fig:hero}
\vspace{0.5em}
\end{strip}

\begin{abstract}
Rendering 3D scenes as pixel art requires that discrete pixels remain stable as the camera moves.
Existing methods snap the camera to a grid. Under orthographic projection, this works: every pixel shifts by the same amount, and a single snap corrects all of them. Perspective breaks this. Pixels at different depths drift at different rates, and no single snap corrects all depths.
Texel splatting avoids this entirely.
Scene geometry is rendered into a cubemap from a fixed point in the world, and each texel is splatted to the screen as a world-space quad.
Cubemap indexing gives rotation invariance. Grid-snapping the origin gives translation invariance.
The primary limitation is that a fixed origin cannot see all geometry; disocclusion at probe boundaries remains an open tradeoff.
\end{abstract}

%% ============================================================
\section{Introduction}
\label{sec:intro}

% Overview
Texel splatting is a rendering technique for perspective-stable pixel art.
Scene geometry is rendered into a cubemap from a fixed probe origin, shaded per cubemap texel, and splatted to the screen as world-space quads.

% Orthographic pixel art works
Orthographic pixel art works because the projection is linear: a world-space offset produces the same screen-space displacement regardless of depth.
Camera movement shifts every pixel by the same amount.
Snapping the camera to the screen grid~\cite{t3ssel8r, propixelizer} compensates for this shift uniformly, and pixels stay locked.

% Perspective breaks this
Perspective breaks this.
Under perspective projection, the same world-space offset produces different screen-space displacements at different depths.
Near geometry displaces more on screen than far geometry.
No single grid snap compensates for all depths simultaneously, and no screen-space correction can.
ProPixelizer documents this effect, called shimmer or pixel creep: ``pixel creep can only ever be solved for orthographic projections''~\cite{propixelizer}.

% Texel marching (concurrent work)
Texel marching~\cite{tesseractcat} addresses the same problem using cubemaps from fixed origins with depth-based reprojection.
Rays are marched through the cubemap in screen space, sampling stored color when intersections are found.
Both methods share the same principle: cubemap parameterization from a fixed origin decouples texel stability from camera movement.
The display methods differ: texel marching raymarches in screen space, texel splatting projects world-space quads.

% Cubemaps
Cubemaps~\cite{greene1986} index by direction, not screen position.
They are rotation-invariant by construction.
Scene geometry is rasterized into cubemaps~\cite{saito1990} and shaded per cubemap texel.
Lighting happens in cubemap space, not screen space, so the results are camera-independent.

% Probes
A cubemap needs an origin.
If the origin moves with the camera, texel assignments shift frame-to-frame.
Snapping the origin to a world grid~\cite{engel2007} gives translation invariance.
The probe origin is the camera position rounded to the nearest grid vertex.

% Splatting
Each visible cubemap texel becomes a world-space quad.

% Contribution
The architecture guarantees rotation invariance through cubemap parameterization and translation invariance through grid-snapped origins, producing perspective-stable pixel art without screen-space correction.

\begin{figure*}[t]
\centering
\begin{subfigure}[t]{0.32\textwidth}
  \includegraphics[width=\textwidth]{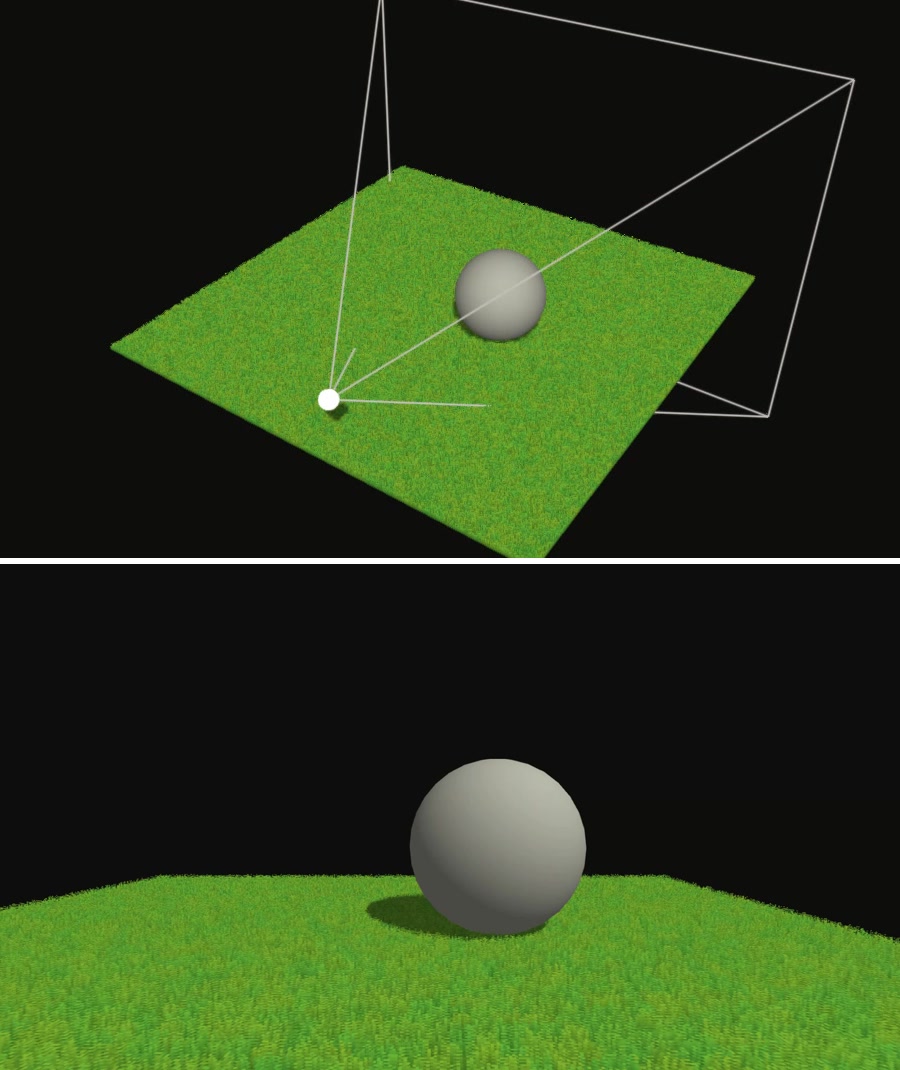}
  \caption{Scene geometry}
  \label{fig:pipeline-raster}
\end{subfigure}
\hfill
\begin{subfigure}[t]{0.32\textwidth}
  \includegraphics[width=\textwidth]{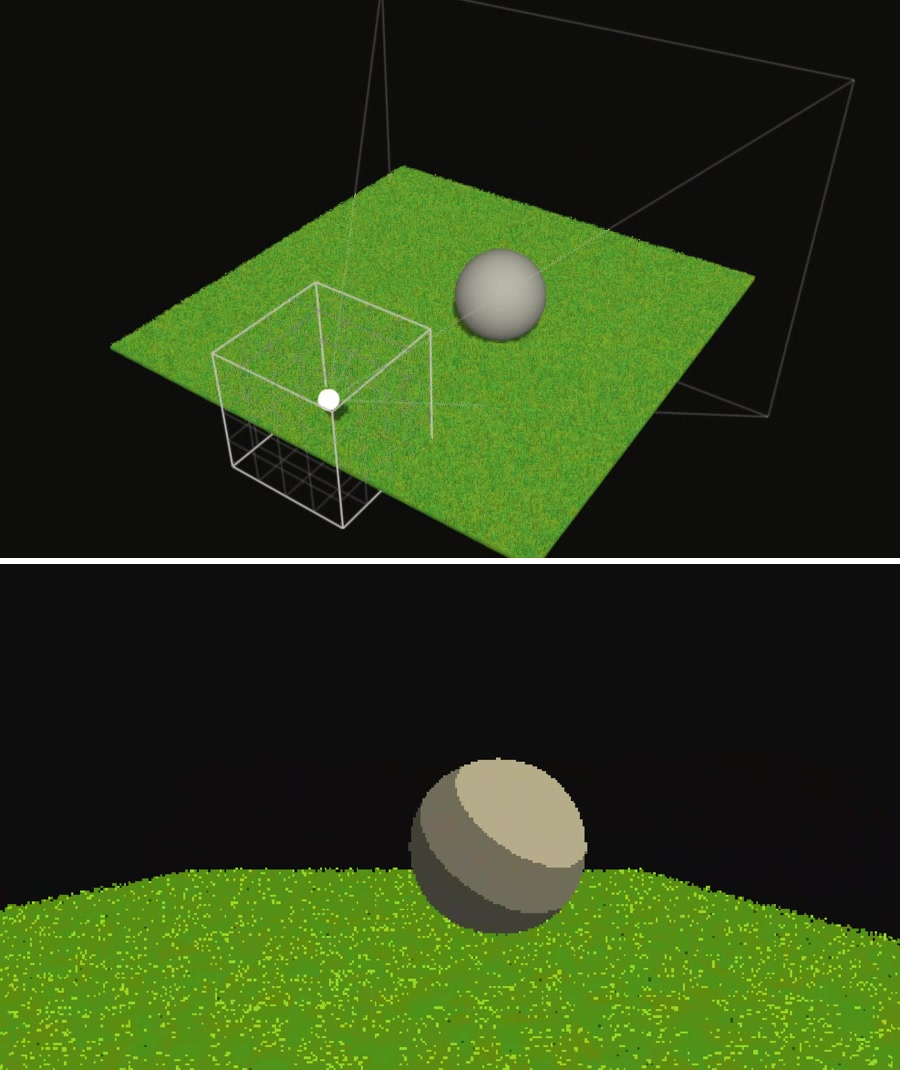}
  \caption{Cubemap capture}
  \label{fig:pipeline-cubemap}
\end{subfigure}
\hfill
\begin{subfigure}[t]{0.32\textwidth}
  \includegraphics[width=\textwidth]{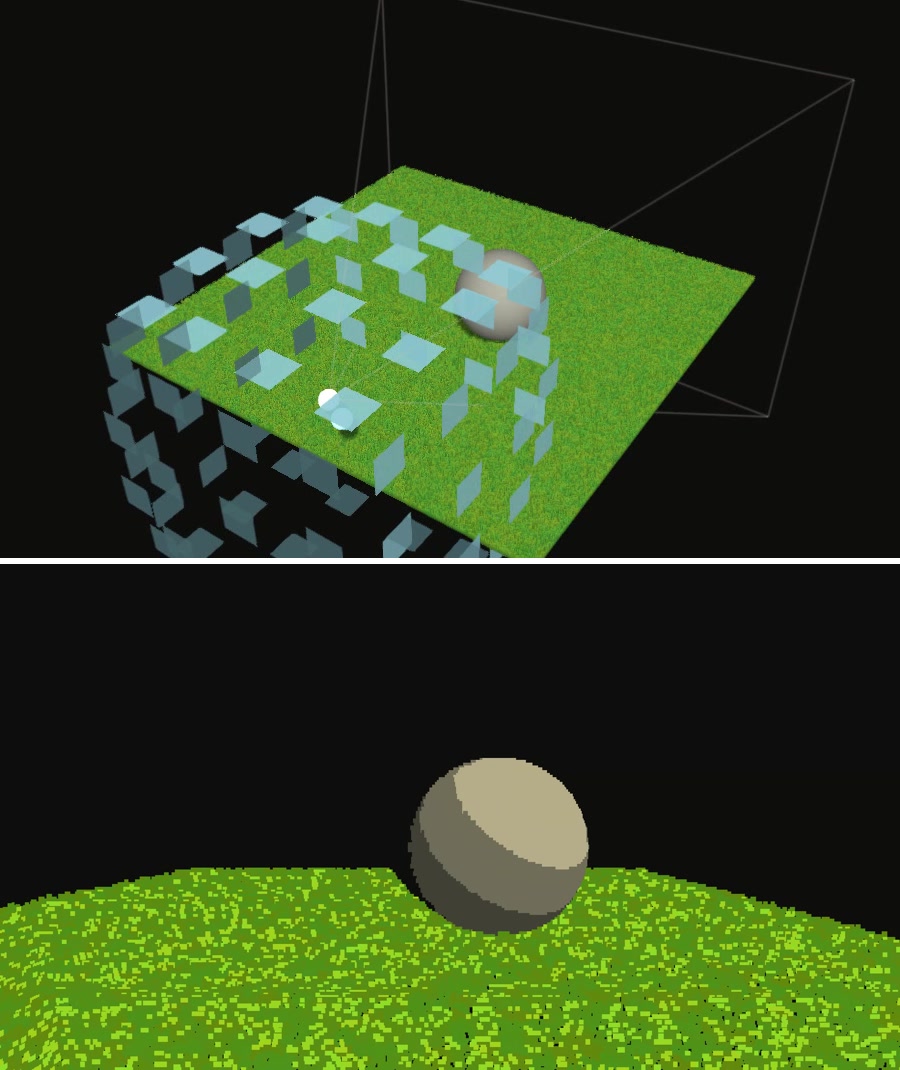}
  \caption{Texel splatting}
  \label{fig:pipeline-splatting}
\end{subfigure}

\caption{Pipeline overview. Each column pairs an orthographic diorama (top) with the corresponding camera view (bottom). The white dot marks the probe origin; the wireframe shows the camera frustum. (a)~Scene geometry rasterized from the probe origin. (b)~Cubemap captured around the probe. (c)~Visible texels splatted as world-space quads.}
\label{fig:pipeline}
\end{figure*}

%% ============================================================
\section{Method}
\label{sec:method}

Scene geometry is rasterized into cubemaps from fixed probe origins, shaded per texel in a compute pass, and splatted to screen as world-space quads (Fig.~\ref{fig:pipeline}).
The pipeline runs each frame: capture, shade, splat.

\subsection{Cubemap capture}
\label{sec:capture}

Scene geometry is rasterized from the probe origin into cubemaps: position, normal, material, and object ID per texel.
Texels are indexed by direction from the origin, giving rotation invariance.

\subsection{Shading}
\label{sec:shading}

A compute pass shades each cubemap texel.
Diffuse lighting from directional and point sources; shadow rays traced against scene geometry.
Shading is camera-independent.
OKLab posterization~\cite{ottosson2020} quantizes lightness to discrete bands, producing flat color steps across surfaces.

Outlines are detected in the same pass.
Object ID discontinuities between neighboring texels mark silhouette edges; normal discontinuities mark creases.
Both are applied as lightness shifts in OKLab space.
Effects applied per texel (posterization, outlines, lighting) inherit this stability; screen-space effects (bloom, haze, light shafts) applied after splatting remain camera-dependent.

\subsection{Probes}
\label{sec:probes}

The probe origin is the camera position snapped to a world grid.
Moving within a grid cell does not change the origin, giving translation invariance.

Three probes are maintained: an eye probe at the camera position, a grid probe at the snapped origin, and a previous probe that holds the old grid cell during transitions.
The eye probe provides disoccluded content (\S\ref{sec:disocclusion}).
The grid probe provides stable content.
The previous probe enables blending between grid cells.

When the camera crosses a cell boundary, the current grid origin becomes the previous origin, and the snapped position becomes the new grid origin.
A $4\times4$ Bayer dither pattern~\cite{bayer1973} crossfades between the two grid probes.
Probe updates are amortized: the eye probe renders every frame; the grid and previous probes alternate, maintaining consistent cost during movement.

\subsection{Splatting}
\label{sec:splatting}

Each visible cubemap texel is splatted to screen as a world-space quad~\cite{pfister2000, zwicker2001}.

Each texel stores a scalar depth $d$, the Chebyshev distance from the probe origin.
The cubemap maps texel coordinates $(u, v)$ on face $F$ to a direction $\mathbf{r}$~\cite{greene1986}.
The world position is
\begin{equation}
  \mathbf{p} = \mathbf{o} + \frac{d}{\|\mathbf{r}\|_\infty}\,\mathbf{r}
  \label{eq:reconstruct}
\end{equation}
where $\mathbf{o}$ is the probe origin and $\|\mathbf{r}\|_\infty = \max(|r_x|, |r_y|, |r_z|)$.
Quad corners evaluate this at $(u \pm h, v \pm h)$ for half-texel width $h$.
All four share depth $d$; no per-corner intersection is needed.

Adjacent texels leave gaps when splatted at exact texel boundaries.
Expanding each quad beyond its boundary fills these gaps, but where neighboring texels lie at similar depths, overlapping quads produce Z-fighting.
Each texel's depth is compared against its four neighbors.
At edges where depths differ, quads expand freely; the depth buffer resolves overlap.
At edges where depths are similar, expansion is constrained and scaled by the grazing angle between the view direction and the cubemap face.

\subsection{Disocclusion}
\label{sec:disocclusion}

A fixed probe origin cannot see geometry occluded from its position.
As the camera moves away from the probe, regions become visible that the probe never captured.
An eye probe at the camera position fills these gaps.
Because the eye probe moves with the camera, its texel assignments are unstable: the same surface point maps to different texels each frame, producing shimmer~\cite{benard2011}.

%% ============================================================
\section{Discussion}
\label{sec:discussion}

\subsection{Cost model}
\label{sec:cost}

The scene in Fig.~\ref{fig:hero} renders at 40\,fps on an iPhone~15 (A16 GPU, mobile Safari) and under 4\,ms per frame on an RTX~4090, both at $384^2$ texels per face.
The pixel art aesthetic demands low cubemap resolution, so the shading budget is inherently small.
Camera-independent shading enables caching across frames~\cite{burns2010, hillesland2016, cook1987}.

\subsection{Limitations and future work}
\label{sec:limitations}

Disocclusion is the primary limitation.
Geometry hidden from the probe origin was never captured; the eye probe fills these gaps, but shimmer is confined to regions the grid probe cannot see.

Probe density controls the balance: large cells produce more stable content but increase the disoccluded area; small cells reduce disocclusion at the cost of more frequent transitions.
Adaptive probe placement, additional stable probes, specular materials, animated geometry, and level-of-detail across probe cells remain open.

%% ============================================================

\bibliographystyle{IEEEtran}

\end{document}